\documentclass{article}
\usepackage{times}

\usepackage{iclr2027_conference,times}

\usepackage[accsupp]{axessibility}
\usepackage{xcolor}
\usepackage{colortbl} 
\usepackage{booktabs}
\usepackage{graphicx}
\usepackage{multirow}
\usepackage{makecell}
\usepackage{tabularx}
\usepackage{comment}
\usepackage{amssymb}
\usepackage{caption}
\usepackage{wrapfig}
\usepackage{enumitem}

\definecolor{mycolor}{RGB}{235,242,250}
\definecolor{variantgray}{gray}{0.42}

\usepackage{amsmath,amsfonts,bm}

\def\eqref#1{equation~\ref{#1}}

\def\1{\bm{1}}

\DeclareMathAlphabet{\mathsfit}{\encodingdefault}{\sfdefault}{m}{sl}
\SetMathAlphabet{\mathsfit}{bold}{\encodingdefault}{\sfdefault}{bx}{n}

\usepackage{hyperref}
\usepackage{url}

\title{Lifelong Small-Object Navigation in Changing Object Layouts: A Benchmark and Method}

\author{
\textbf{Jiagan~Huang},
\textbf{Zikun~Zhou}\textsuperscript{*},
\textbf{Zijian~Ni},
\textbf{Hongpeng~Wang},\\
\textbf{Guangming~Lu},
\textbf{Jun~Yu},
and \textbf{Wenjie~Pei}\textsuperscript{*}\\[3pt]
Harbin Institute of Technology, Shenzhen\\[-1pt]
\textsuperscript{*}\small Corresponding authors
}
\iclrfinalcopy
\begin{document}

\maketitle

\begin{abstract}
Household robots need to continually navigate to different objects in the same environment, many of which are small and portable, such as tools and toys. Their small visual footprint and frequent occlusion make reliable observation difficult, and they may be moved by people without the robot observing the changes. We formulate this challenging task as Lifelong Small-object Navigation in Changing Object Layouts (\textbf{LiSoNav-COL}). Agents must seek suitable viewpoints for reliable observation, accumulate and reuse scene knowledge to efficiently locate subsequent targets, and update outdated memory after object relocation. To eliminate the need for prior scene scanning, we also require agents to start navigation with empty scene memory. Although practical, this task still lacks benchmarks designed around its defining assumptions. To bridge this gap, we introduce \textbf{LiSoNav-Eval}, a dedicated benchmark spanning 28 indoor scenes with 45 small-object categories. Its lifelong navigation sequences include both unchanged and relocated targets to evaluate memory reuse and adaptation to object relocation. To address this challenging task, we propose a navigation method based on multi-view Inspection with Viewpoint-Anchored Memory, dubbed \textbf{IVAM-Nav}. IVAM-Nav actively observes supporting surfaces from complementary viewpoints for reliable small-object perception and anchors the resulting memory to their observation viewpoints, supporting relational memory reuse and revalidation under similar viewing conditions. 
Extensive experiments on LiSoNav-Eval demonstrate favorable performance of IVAM-Nav against representative methods. Benchmark analyses also show that smaller objects, larger environments, and longer relocation distances pose greater challenges. The dataset and code are available \href{https://github.com/hhhhhjg/LiSoNav-Benchmark/tree/main}{here}.

\end{abstract}
\begin{figure*}[h]
    \vspace{-4mm}
    \centering
    \includegraphics[width=\textwidth]{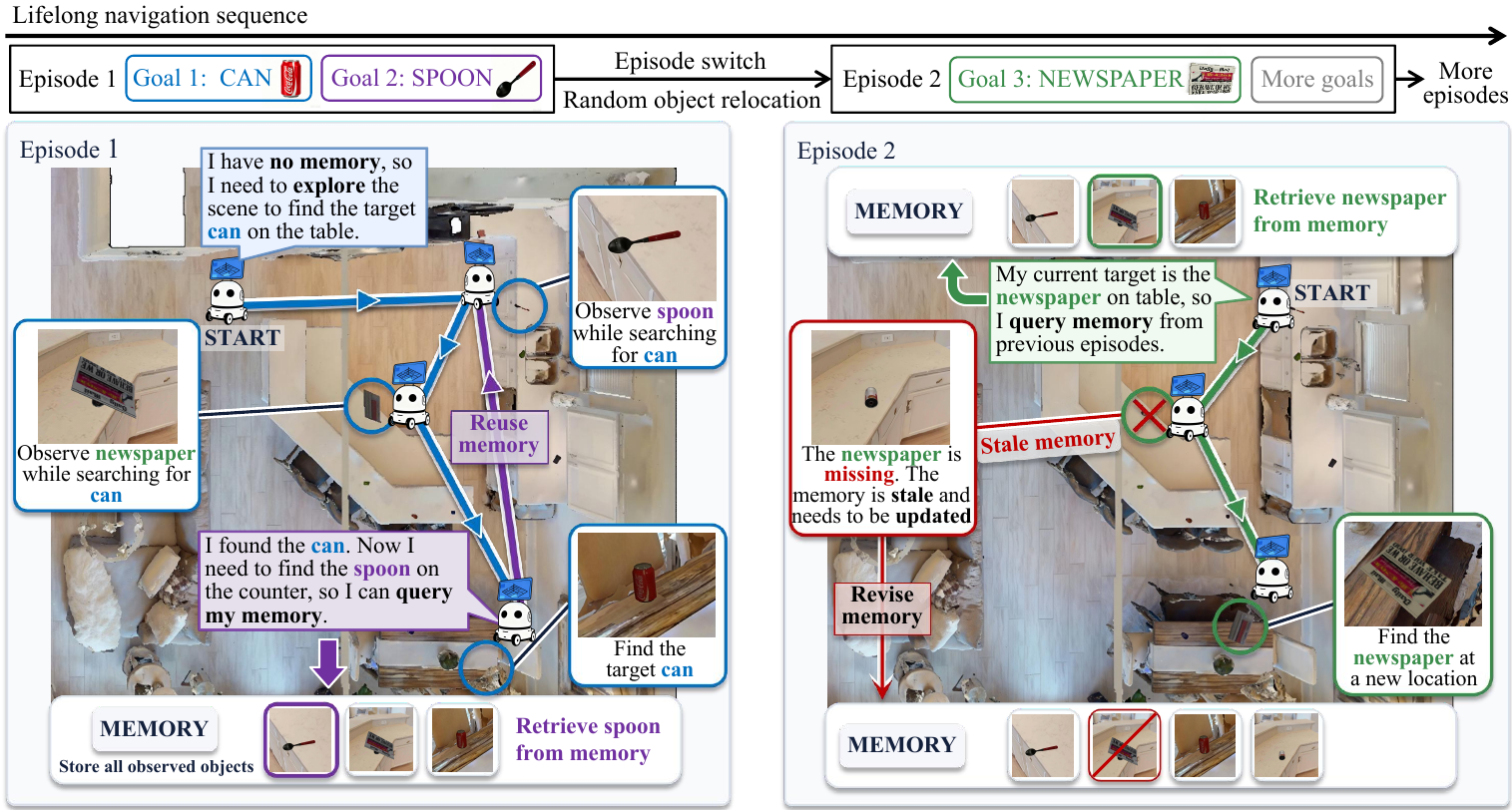}
    \vspace{-5mm}
\caption{\textbf{An example navigation sequence in LiSoNav-Eval,} in which the agent is instructed to find the \textcolor[RGB]{0,112,192}{CAN} on the table, the \textcolor[RGB]{112,48,160}{SPOON} on the counter, the \textcolor[RGB]{58,141,75}{NEWSPAPER} on the table, and other targets in sequence. Starting with no prior scene memory, the agent searches for the \textcolor[RGB]{0,112,192}{CAN} while constructing memory from observed objects, which is later reused to find the \textcolor[RGB]{112,48,160}{SPOON}. Between episodes, randomly selected objects are relocated without the agent observing the changes. When searching for the \textcolor[RGB]{58,141,75}{NEWSPAPER}, the agent first revisits the remembered location, finds the newspaper missing, updates the scene memory, and then locates the newspaper elsewhere. This example illustrates how LiSoNav-Eval evaluates small-object navigation as agents accumulate scene knowledge, reuse it across tasks, and respond to changes in object locations.}
    \label{fig:reslon_overview}
\vspace{-5mm}
\end{figure*}

\vspace{-2mm}
\section{Introduction}
\vspace{-2mm}

Object-goal navigation requires embodied agents to search for and navigate to target objects specified by images or language~\citep{krantz2023navigating,zhu2021soon}. In household environments, robots may repeatedly perform such tasks in the same scene, motivating lifelong navigation that retains and reuses experience across successive tasks~\citep{khanna2024goatbench}. Scene memory supports this process by preserving previously acquired object information to reduce redundant exploration.

In everyday household activities, many objects that robots need to locate and manipulate are small, portable items, such as tools and toys. People may move objects unnoticed by the robot, leaving its scene memory outdated and requiring verification and updates. These objects also occupy only a small portion of the visual field and are often occluded, making them difficult to observe clearly~\citep{wang2023smallobject}. 
These real-world challenges motivate us to study \textbf{Li}felong \textbf{S}mall-\textbf{o}bject \textbf{Nav}igation in \textbf{C}hanging \textbf{O}bject \textbf{L}ayouts (\textbf{LiSoNav-COL}). In this task, an agent starts with no prior scene memory and navigates to successive small-object goals within the same environment, incrementally acquiring scene knowledge throughout the navigation. Given that unobserved object relocations may render its memory outdated, the agent must learn to reuse accumulated scene knowledge while adapting to these changes for accurate and efficient navigation. Starting with an empty memory eliminates the need for a separate scene-scanning phase in real-world applications.

Existing navigation benchmarks adopt task assumptions that differ from those of LiSoNav-COL. The mainstream lifelong navigation benchmarks~\citep{wani2020multion,khanna2024goatbench,Wang2026ExploreWL} mainly evaluate experience reuse in static scenes, where target locations remain unchanged across subtasks. OpenIN~\citep{tang2025openin} evaluates sequential navigation after object relocation, but starts with a preconstructed scene graph and relocates objects before the task sequence begins. This setting assesses the use of prior scene knowledge after relocation, rather than its online acquisition and maintenance through repeated changes. Meanwhile, the small-object navigation benchmark~\citep{wang2023smallobject} focuses on perception and target search in static scenes under a single-goal setting. These evaluation settings do not jointly capture the requirements of LiSoNav-COL: acquiring scene knowledge from an initially empty memory, reusing it across small-object navigation tasks, and adapting to repeated object relocation.

To support systematic evaluation of LiSoNav-COL, we introduce \textbf{LiSoNav-Eval}, a new benchmark comprising navigation sequences and an evaluation protocol for this task.
Figure~\ref{fig:reslon_overview} presents an illustrative example of our benchmark, where the agent completes
a sequence of small-object navigation subtasks (each corresponding to a specific goal) in the same environment, starting with no prior scene memory and retaining the knowledge acquired during task execution. Specifically, each navigation sequence comprises multiple episodes, each containing several navigation subtasks. Object locations remain unchanged within each episode, while randomly selected objects are relocated between episodes without the agent observing the changes. 
By including both unchanged and relocated targets, LiSoNav-Eval preserves opportunities to benefit from previously acquired knowledge while testing adaptation when remembered locations become outdated. Together, these settings support the evaluation of small-object navigation as agents accumulate scene knowledge, reuse it across tasks, and respond to changes in object layouts.

In LiSoNav-COL, incomplete or unreliable small-object observations can limit the effectiveness of existing scene-memory methods~\citep{huang2025msgnav,yang2025threedmem,tang2025openin}. Object relocation compounds these observation difficulties, making memory reuse and revalidation more challenging.
In this paper, we propose \textbf{IVAM-Nav}, a lifelong small-object navigation
method based on multi-view \textbf{I}nspection and
\textbf{V}iewpoint-\textbf{A}nchored \textbf{M}emory, to address these challenges. Our key idea is to actively inspect the supporting surfaces where the target may be located from multiple viewpoints and anchor the memory to these viewpoints.
Specifically, the agent acquires reliable and complementary observations around potential supporting surfaces and stores them with their camera poses and semantic associations. These records enable reobservation under similar viewing conditions to effectively verify and update the memory.
Extensive experiments on LiSoNav-Eval demonstrate the effectiveness of our algorithm. Our contributions can be summarized as:

\vspace{-1mm}
\begin{itemize}[leftmargin=1.5em]
\item We formulate LiSoNav-COL, a new task for lifelong small-object navigation from empty scene memory under changing object layouts. We also introduce LiSoNav-Eval, a dedicated benchmark for evaluating agent performance on this task.

\item We propose IVAM-Nav, which constructs a viewpoint-anchored graph memory through active multi-view inspection. This design supports relational memory reuse and memory revalidation under similar viewing conditions, enabling robust navigation.

\item Extensive experiments on LiSoNav-Eval demonstrate favorable performance of IVAM-Nav against representative methods, and benchmark analyses identify smaller objects, larger environments, and longer relocation distances as persistent challenges.
\end{itemize}
\vspace{-1mm}
\section{Related Work}
\vspace{-2mm}

\noindent\textbf{Lifelong navigation in changing environments.}
Existing navigation tasks and benchmarks span single-goal, sequential, and lifelong settings in largely static scenes~\citep{chaplot2020objectnav,krantz2023navigating,zhu2021soon,wani2020multion,khanna2024goatbench,miao2026langmap,yu2025cnav,wang2026lenl}, while dynamic settings consider moving objects and spatiotemporal retrieval~\citep{kurenkov2023modeling,dorbala2026personalized,jiang2025dualmap,bogenberger2026glasses,chen2025star}. Most closely related to our setting, OpenIN studies changing object layouts but requires a preconstructed scene representation before navigation, with its evaluation data and implementation currently unavailable~\citep{tang2025openin}. LiSoNav-Eval instead evaluates lifelong small-object navigation from empty memory under changing object layouts.
\vspace{-0.5mm}

\noindent\textbf{Memory construction, reuse, and revalidation.}
Existing navigation methods construct and query scene memory through object- or scene-centric graphs, visual observations, and spatial representations~\citep{gu2024conceptgraphs,werby2024hovsg,yang2025threedmem,liu2025dynamem,jiang2025dualmap,zhu2025mtu3d,chang2024goat,wang2023smallobject,xu2026dream,ge2026evomemnav}, while dynamic-memory methods further update or reassess stored information based on new observations~\citep{tang2025openin,jiang2025dualmap,liu2025dynamem,yan2025dovsg,zhao2026supermap,ko2025care}. APRR and MSGNav actively adjust viewpoints after target detection or localization for manipulation or final-view selection~\citep{qin2025aprr,huang2025msgnav}. In contrast, IVAM-Nav builds reliable small-object memory via complementary multi-view inspection and uses acquisition viewpoints as persistent anchors for relational retrieval and revalidation.

\vspace{-1mm}
\section{LiSoNav-Eval Benchmark}
\label{sec:lison_eval}

\subsection{Task Definition}
\label{sec:lison_cs_task}
\vspace{-2mm}
The agent is required to complete a sequence of navigation subtasks (each corresponding to a small-object goal) in the same indoor environment with no prior scene memory. It needs to retain the knowledge acquired during subtask execution to facilitate the subsequent subtasks. The lifelong navigation sequence in LiSoNav-Eval spans three or four consecutive episodes, with each episode containing two or three navigation subtasks. Figure~\ref{fig:reslon_overview} illustrates such a sample.

The goal object of each subtask is specified by a natural language instruction. To clearly specify the target object, the instruction is designed to include coarse contextual cues, such as its \textit{supporting surface} or \textit{nearby surroundings}, while leaving its exact current location unspecified. This formulation also aligns with how people typically describe the location of a small object in a home environment. Herein, we show two instruction samples:

\vspace{-1mm}
\begin{itemize}
    \item \textbf{Instruction 1:} \textit{Find the tissue box on the speckled countertop near the sink.}
\vspace{-1mm}

    \item \textbf{Instruction 2:} \textit{Find the turquoise Lego Duplo brick on a small wooden table next to a bed.}
\end{itemize}
\vspace{-1mm}

We also provide an optional reference image to aid small-object recognition, which reveals no information about the current location of the target. In real-world applications, such images could be obtained through online searches or other external sources.
Within an episode, object locations remain unchanged, and subtasks are executed sequentially from the terminal pose of the preceding subtask. Objects are relocated only between episodes to render memory outdated, not during navigation, to avoid turning the task into online tracking of moving objects.

A subtask is successful when the agent stops within 1.5\,m of the valid instance matching the target description. Each subtask has a path-length budget of 10 times the shortest geodesic distance from the start to the target, and terminates immediately once this budget is exceeded. This budget allows exploration and recovery from outdated location information after object relocation while limiting excessive search. We report standard navigation metrics (SR and SPL), together with Historical Revisit (HR) and Relocated-target SR. Detailed definitions are provided in~\ref{sec:exp_setup}.

\begin{table*}[t]
\vspace{-2mm}
\centering
\setlength{\tabcolsep}{4pt}
\renewcommand{\arraystretch}{1.1}
\caption{Comparison of LiSoNav-Eval with existing lifelong navigation benchmarks. \textsuperscript{\(\dagger\)} indicates that the corresponding evaluation data are not publicly released.}
\vspace{-3mm}
\resizebox{\textwidth}{!}{
\begin{tabular}{lccccccccc}
\toprule
& Scenes &
Subtasks &
\makecell{Semantic target\\categories} &
\makecell{Object\\relocation} &
\makecell{Relocated\\subtasks} &
\makecell{Relocated\\ratio} &
\makecell{Prior scene\\knowledge} &
\makecell{Subtasks\\per sequence} \\
\midrule

MultiON 3-ON val~\citep{wani2020multion}
& 11 & 3,150 & 1 & No & -- & -- & No & 3.00 \\

GOAT val-unseen~\citep{khanna2024goatbench}
& 36 & 2,669 & 36 & No & -- & -- & No & 7.41 \\

OpenIN\textsuperscript{\(\dagger\)}~\citep{tang2025openin}
& 4 & -- & -- & Yes & -- & 100\% & Yes & 4--5 \\

LMEE-Bench test~\citep{Wang2026ExploreWL}
& 36 & 828 & 38 & No & -- & -- & No & 4.99 \\

\rowcolor{mycolor}
\textbf{LiSoNav-Eval} (Ours)
& 28 & 1,199 & 45 & Yes & 545 & 45.45\% & No & 8.27 \\

\bottomrule
\end{tabular}}
\label{tab:benchmark_comparison}
\vspace{-5mm}
\end{table*}

\vspace{-1mm}
\subsection{LiSoNav-Eval Dataset}
\label{sec:lison_eval_dataset}
\vspace{-2mm}
\noindent\textbf{Scene and object assets.}
We use indoor scenes from HM3D~\citep{ramakrishnan2021hm3d} and select 82 diverse small-object instances across 45 categories from YCB~\citep{calli2015ycb} and HSSD~\citep{khanna2024hssd}, covering portable household items such as cups, balls, tools, and fruits.

\noindent\textbf{Dataset construction.}
To construct a lifelong navigation sequence, we first populate an HM3D scene with small objects. We use Qwen-VL-Max~\citep{bai2023qwenvl} to identify suitable supporting surfaces and define category-specific placement rules. These rules exclude implausible object-surface combinations, such as placing food on beds. 
We also limit local object density to avoid excessive clutter and occlusion that would make small objects unnecessarily difficult to observe. Based on these rules, we randomly generate an initial object configuration for each scene and sample the target objects episode by episode. Targets selected in the first episode remain at their initial locations. In each subsequent episode, a selected target has a probability of being relocated to another compatible supporting surface, preserving plausible object placements after relocation.
The same target is not selected more than once within an episode, but may reappear in later episodes, allowing previously acquired knowledge to be useful for subsequent subtasks.

Since LiSoNav-COL starts from empty scene memory, relocating an object that was not previously observed may not invalidate any acquired memory. To increase the likelihood that relocation creates stale memory, we preferentially sample objects in the same semantic region as, and within $5\,\mathrm{m}$ of, a previously selected target, as they are more likely to have been observed during earlier navigation.

\noindent\textbf{Dataset verification.}
To ensure each target is visually identifiable from a suitable viewpoint, we sample observations around every target with the target near the view center. We use Qwen-VL-Max to assess target recognizability and manually remove overly difficult cases.

\noindent\textbf{Dataset annotation.}
We provide a textual description for each of the 82 small-object instances, emphasizing attributes that distinguish visually similar instances. For coarse contextual cues, we use a real observation around the target location and Qwen-VL-Max to generate coarse descriptions of the supporting surface and available surroundings using attributes such as category, color, and material, while excluding fine-grained spatial details that could reveal the target's precise location.

\noindent\textbf{Dataset statistics.}
As shown in Table~\ref{tab:benchmark_comparison}, LiSoNav-Eval contains 145 lifelong navigation sequences from 28 HM3D scenes, comprising 500 episodes and 1,199 subtasks across 45 categories and 82 small-object instances. 545 (45.45\%) of these subtasks involve relocated targets. Each navigation sequence contains 8.27 subtasks on average. Compared with OpenIN~\citep{tang2025openin}, LiSoNav-Eval spans more scenes and includes more subtasks per sequence. Besides, LiSoNav-Eval also supports evaluation of both knowledge reuse and adaptation to object relocation.

\begin{figure*}[t]
\vspace{-4mm}
    \centering
    \includegraphics[width=\textwidth]{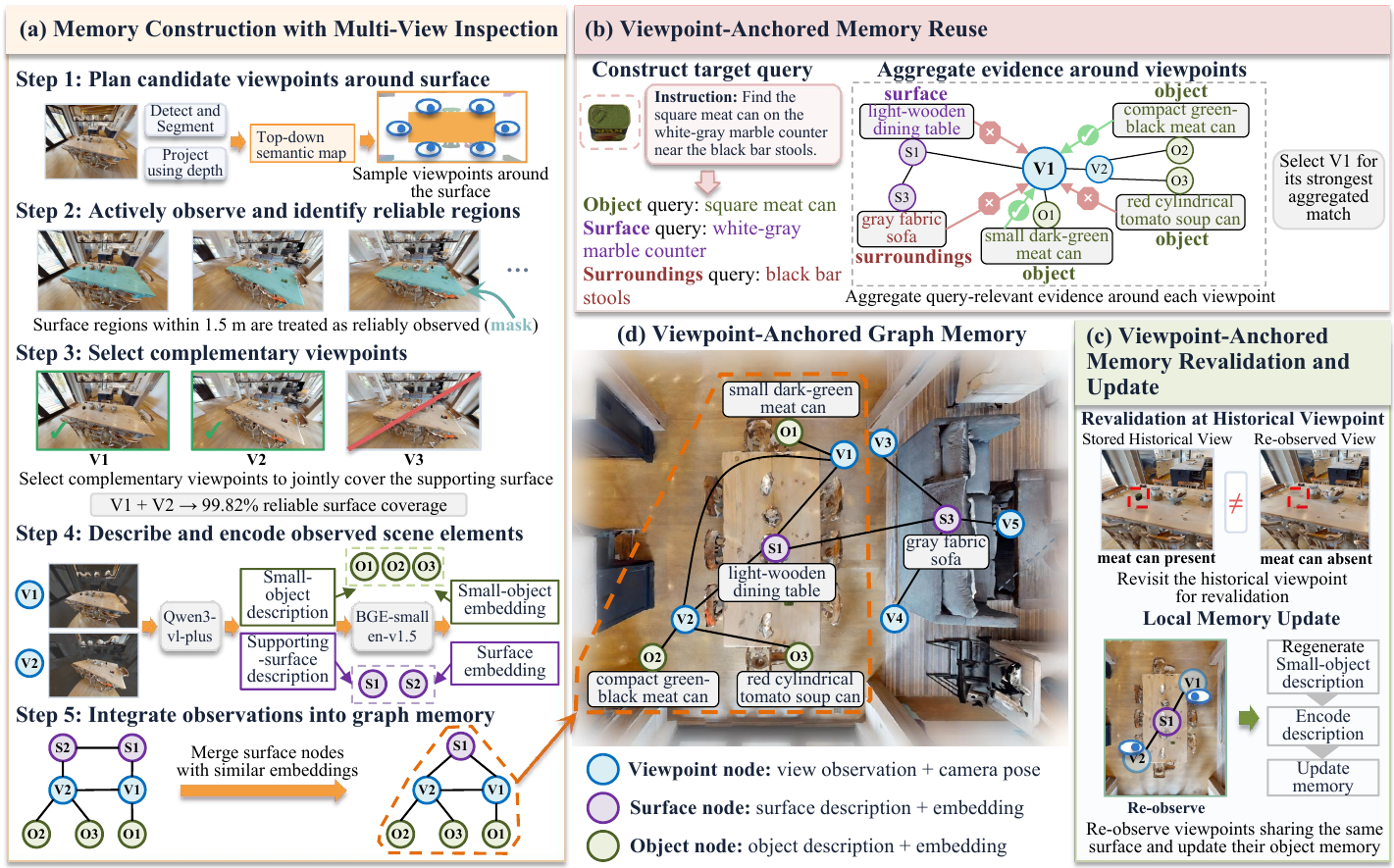}
    \vspace{-5mm}
    \caption{\textbf{Overview of IVAM-Nav.}
IVAM-Nav maintains a viewpoint-anchored graph memory through three processes:
(a) \textbf{Memory construction with multi-view inspection} actively observes supporting surfaces from complementary viewpoints to acquire reliable small-object evidence.
(b) \textbf{Viewpoint-anchored memory reuse} aggregates relational evidence around historical viewpoints to retrieve those relevant to the current query.
(c) \textbf{Viewpoint-anchored memory revalidation and update} revisits retrieved historical viewpoints to verify memory under similar viewing conditions and locally update inconsistent information.}
\label{fig:method_overview}
\vspace{-4mm}
\end{figure*}

\vspace{-1mm}
\section{IVAM-Nav}
\label{sec:method}
\vspace{-2mm}
To handle this challenging task, we propose IVAM-Nav, which is designed to actively inspect supporting surfaces from multiple viewpoints for reliable small-object perception and anchor its memory to these viewpoints. For the first task in a new environment, the agent searches for and navigates to a supporting surface where the target may be located. It then samples nearby viewpoints and retains a small complementary set to obtain clear observations of the supporting surface. These observations, together with their poses and related scene information, form a viewpoint-anchored graph memory. 
For subsequent subtasks, the agent aggregates the small-object and supporting-surface memories attached to each historical viewpoint and evaluates their relevance to the target query to retrieve the most relevant viewpoint. This design enables the agent to revisit previously recorded viewpoints and verify memory under similar viewing conditions.

\vspace{-1mm}
\subsection{Overall Framework}
\label{sec:method_overview}
\vspace{-2mm}

IVAM-Nav is a navigation framework built on a viewpoint-anchored graph memory. It comprises three core processes: (a) memory construction with multi-view inspection, (b) viewpoint-anchored memory reuse, and (c) viewpoint-anchored memory revalidation and update, as shown in Figure~\ref{fig:method_overview}.

For each navigation subtask, the agent first queries the memory and revisits relevant historical viewpoints. At each revisited viewpoint, it checks the current observation for the target and compares it with the stored observation to assess memory validity. Detected inconsistencies trigger reobservation and updates at viewpoints associated with the same supporting surface. When useful historical candidates are unavailable or exhausted, the agent explores unobserved regions and actively inspects selected supporting surfaces, constructing new memories for subsequent tasks. The agent stops once the MLLM confirms the target in a current observation. Next we detail the three core processes.

\vspace{-1mm}
\subsection{Persistent Viewpoint-Anchored Graph Memory}
\vspace{-2mm}
The memory of IVAM-Nav is initialized empty and progressively accumulated throughout the entire lifelong navigation sequence. As shown in Figure~\ref{fig:method_overview}(d), it is organized as a viewpoint-anchored graph $\mathcal{G}=(\mathcal{V},\mathcal{E})$ with \textit{viewpoint}, \textit{surface}, and \textit{object} nodes. Viewpoint nodes store observations and camera poses, while surface and object nodes store semantic descriptions and embeddings and are connected to their observation viewpoints. Viewpoints observing the same surface are connected, while nearby distinct surfaces are linked, enabling relational aggregation during retrieval. Historical viewpoints thus serve as anchors for memory retrieval and preserve the poses needed for revalidation under similar viewing conditions.

\vspace{-1mm}
\subsection{Memory Construction with Multi-View Inspection}
\vspace{-2mm}
To construct reliable memory, IVAM-Nav combines frontier-based exploration with active multi-view inspection of selected supporting surfaces, as illustrated in Figure~\ref{fig:method_overview}(a). During exploration, RGB-D observations are used to detect and segment supporting surfaces, which are projected onto a top-down semantic map using depth and camera poses. Based on the target instruction, an MLLM selects either a frontier for further exploration or a nearby supporting surface for multi-view inspection. If a surface is selected, the agent samples viewpoints around it based on the semantic map (Step~1).
The agent then visits the sampled viewpoints to acquire close-range observations (Step~2). For each viewpoint $v$, we define its reliable observation region $\mathcal{R}(v)$ as the visible surface region within a predefined observation range, visualized by the mask in Step~2 of Figure~\ref{fig:method_overview}(a). We then select a compact complementary subset that jointly covers the supporting surface (Step~3) by greedily maximizing the additional reliable surface coverage:
\begin{equation}
v^{*}=\arg\max_{v\in\mathcal{C}\setminus\mathcal{S}} \frac{\left|\mathcal{R}(v)\setminus\bigcup_{u\in\mathcal{S}}\mathcal{R}(u)\right|}{|\mathcal{R}_{\mathrm{surf}}|},
\end{equation}
where $\mathcal{C}$ and $\mathcal{S}$ denote the visited and selected viewpoint sets, respectively, $\mathcal{R}_{\mathrm{surf}}$ is the supporting-surface region, and $|\cdot|$ denotes the area of a region. Selection stops once the required coverage is reached, reducing redundant observations while preserving reliable small-object information across the supporting surface. For each selected viewpoint, we mask regions beyond the reliable observation range in the corresponding view. We then use an MLLM to describe the visible small objects and supporting surface, and encode the resulting descriptions into semantic embeddings (Step~4). These scene elements are associated with their observation viewpoints and integrated into the graph (Step~5), with semantically consistent observations of the same supporting surface merged into a shared surface node.

\vspace{-1mm}
\subsection{Viewpoint-Anchored Memory Reuse}
\vspace{-2mm}
Memory reuse allows accumulated scene knowledge to guide subsequent navigation, but noisy or incomplete small-object information can make direct object-level retrieval unreliable. IVAM-Nav therefore treats historical viewpoints as retrieval anchors and evaluates them using the object, surface, and surroundings information associated with each viewpoint, as shown in Figure~\ref{fig:method_overview}(b). The instruction is decomposed into object, surface, and surroundings queries and encoded into the same semantic space as the corresponding memory information.

For each viewpoint node, we aggregate query-relevant information within its two-hop graph neighborhood, capturing sufficient local context while limiting unrelated information. The object and surface queries are matched with directly connected object and surface nodes, respectively, such as O1 and S1 for V1 in the figure. The second hop further uses object information from neighboring viewpoints and surroundings information from nearby surfaces. The highest-scoring viewpoints are selected for revisit and memory revalidation.

\vspace{-1mm}
\subsection{Viewpoint-Anchored Memory Revalidation and Update}
\vspace{-2mm}
A retrieved viewpoint provides both a target-search location and an observation reference for memory revalidation. As shown in Figure~\ref{fig:method_overview}(c), IVAM-Nav revisits the viewpoint and uses an MLLM to compare the new observation with the stored view under similar viewing conditions, focusing on small-object presence and spatial arrangement. This reduces viewpoint-induced variation, making inconsistencies more indicative of actual scene changes.
When an inconsistency is detected, IVAM-Nav follows graph relations to re-observe related viewpoints sharing the same surface and updates their small-object descriptions and embeddings. This relation-guided revision removes stale local information without unnecessary global memory reconstruction.

\section{Experiments}
\vspace{-2mm}

\subsection{Experimental Setup}
\label{sec:exp_setup}
\vspace{-2mm}

\noindent\textbf{Baselines.} We select representative baselines with different cross-task memory capabilities. VLFM~\citep{yokoyama2024vlfm} provides a reference without cross-task memory. AstraNav~\citep{hu2026astranav} and 3D-Mem~\citep{yang2025threedmem} retain and reuse scene knowledge across tasks but do not explicitly revise stale information, whereas DualMap~\citep{jiang2025dualmap} and DynaMem~\citep{liu2025dynamem} additionally incorporate mechanisms for adapting stored scene information to environmental changes. Together, these methods provide comparisons across navigation without cross-task memory, cross-task memory reuse, and memory adaptation under scene changes.

For a consistent task interface, all methods receive the same navigation instructions while retaining their core memory mechanisms. Since DualMap assumes a preconstructed scene map, we initialize it by traversing each environment along a predefined route before evaluation. We additionally report oracle variants of 3D-Mem and DynaMem with ground-truth small-object detections within 1\,m. These variants reduce failures caused by small-object detection, allowing us to examine whether performance differences persist when this perceptual difficulty is partially alleviated. Details of baseline adaptations are provided in Appendix.

\noindent\textbf{Evaluation metrics.}
We report Success Rate (SR) and Success weighted by Path Length (SPL) for overall navigation performance. To evaluate historical-memory retrieval, we additionally introduce \emph{Historical Revisit} (HR) as a behavioral measure. Among subtasks with valid historical target observations, HR is the fraction of such subtasks for which the agent revisits within 1.5 m of a location where the target was previously observed. Whether a target was previously observed is determined using ground-truth simulator information, with detailed criteria provided in Appendix. To evaluate adaptation to object relocation, we further report \emph{Relocated-target SR}, computed over subtasks whose targets have been relocated before the current task.

\noindent\textbf{Implementation details.} IVAM-Nav uses YOLO-World~\citep{cheng2024yoloworld} and SAM-L~\citep{kirillov2023segment} for supporting-surface perception, Qwen3-VL~\citep{bai2025qwen3vl} for MLLM-based operations, and BGE-small-en-v1.5~\citep{xiao2024cpack} for textual semantic embeddings. We use \texttt{qwen3-vl-plus-2025-12-19} for all MLLM calls. For a controlled comparison, baseline components requiring MLLM inference also use the same model. Hyperparameters are fixed across all experiments, with further implementation details provided in Appendix.

\begin{table*}[t]
\vspace{-2mm}
\centering
\setlength{\tabcolsep}{7pt}
\renewcommand{\arraystretch}{1.1}
\caption{Overall performance on LiSoNav-Eval. Oracle variants use ground-truth detection assistance as described in Sec.~\ref{sec:exp_setup}.}
\vspace{-3mm}
\resizebox{0.975\linewidth}{!}{%
\begin{tabular}{lcccccc}
\toprule
Method
& \makecell{Cross-task\\Memory}
& \makecell{Stale-memory\\Revision}
& \makecell{Overall\\SR$\uparrow$}
& \makecell{Overall\\SPL$\uparrow$}
& HR
& \makecell{Relocated-target\\SR$\uparrow$} \\
\midrule
VLFM~\citep{yokoyama2024vlfm}
    & No  & --  & 12.3\% & 8.8\%  & 27.6\% & 11.7\% \\
AstraNav~\citep{hu2026astranav}
    & Yes & No  & 12.2\% & 9.5\%  & 23.2\% & 9.2\% \\
DualMap~\citep{jiang2025dualmap}
    & Yes & Yes & 13.9\% & 10.2\% & 35.2\% & 10.3\% \\

DynaMem~\citep{liu2025dynamem}
    & Yes & Yes & 18.0\% & 12.7\% & 38.1\% & 14.9\% \\
DynaMem Oracle~\citep{liu2025dynamem}
    & Yes & Yes & 20.1\% & 14.4\% & 39.6\% & 15.8\% \\
3D-Mem~\citep{yang2025threedmem}
    & Yes & No  & 21.9\% & 15.5\% & 34.3\% & 17.2\% \\
3D-Mem Oracle~\citep{yang2025threedmem}
    & Yes & No  & 27.5\% & 18.3\% & 49.2\% & 19.3\% \\
\rowcolor{mycolor}
\textbf{IVAM-Nav}
    & Yes & Yes
    & \textbf{33.4\%}
    & \textbf{18.4\%}
    & \textbf{77.4\%}
    & \textbf{30.5\%} \\
\rowcolor{mycolor}
\textcolor{variantgray}{IVAM-Nav (w/ reference image)}
& \textcolor{variantgray}{Yes}
& \textcolor{variantgray}{Yes}
& \textcolor{variantgray}{36.3\%}
& \textcolor{variantgray}{21.3\%}
& \textcolor{variantgray}{79.6\%}
& \textcolor{variantgray}{30.1\%} \\
\bottomrule
\end{tabular}}
\label{tab:overall_results}
\vspace{-5mm}
\end{table*}

\vspace{-1mm}
\subsection{Overall Performance}
\label{sec:overall_results}
\vspace{-2mm}
As shown in Table~\ref{tab:overall_results}, IVAM-Nav achieves an overall SR of 33.4\% and SPL of 18.4\%, outperforming all baselines in SR and remaining competitive in path efficiency. Notably, 3D-Mem Oracle and DynaMem Oracle achieve 27.5\% and 20.1\% SR despite receiving ground-truth small-object detection assistance, suggesting that alleviating small-object perception difficulty alone does not fully address LiSoNav-COL, where accumulated scene knowledge must also be effectively reused and maintained as object locations change. With an additional reference image, IVAM-Nav further reaches 36.3\% SR and 21.3\% SPL, indicating that improved target appearance information can provide additional benefits.
In addition, IVAM-Nav achieves an HR of 77.4\%, substantially higher than the baselines, indicating that its memory frequently guides the agent back to previously observed target locations. However, high HR alone does not imply success in LiSoNav-COL, as historical locations may become outdated after object relocation and require revalidation. On relocated targets, IVAM-Nav achieves 30.5\% SR, compared with 19.3\% for 3D-Mem Oracle and 15.8\% for DynaMem Oracle. Together, these results show that IVAM-Nav effectively reuses accumulated scene knowledge while remaining effective under object relocation.

\begin{figure*}[t]
\centering

\begin{minipage}[t]{0.54\textwidth}
    \vspace{0pt}
    \centering
    \includegraphics[width=\linewidth]{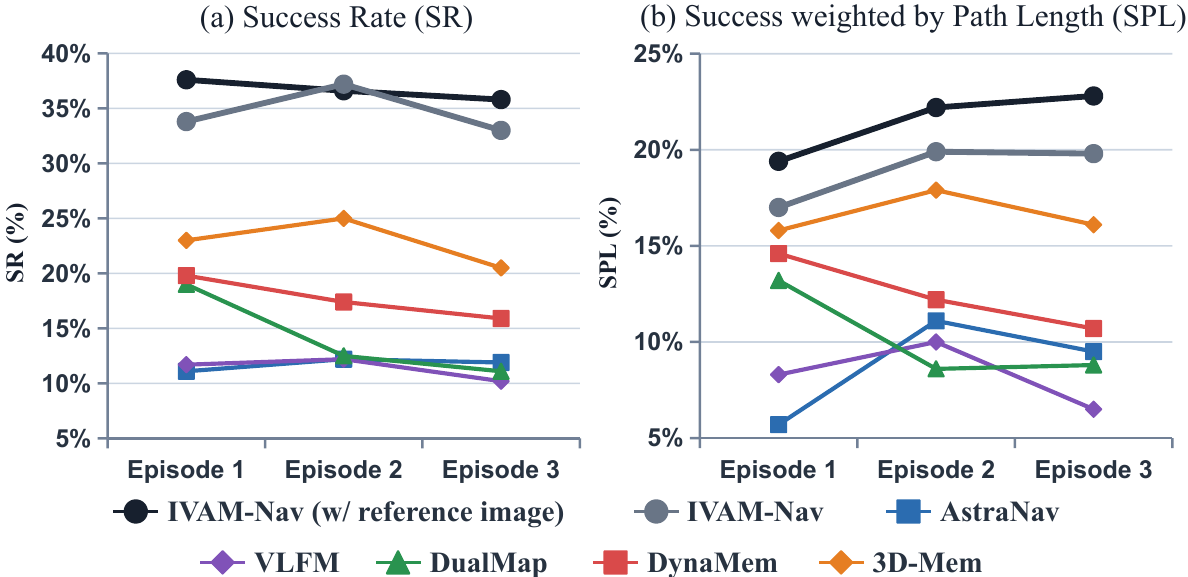}
    \vspace{-6mm}
    \caption{Performance across the first three episodes.}
    \label{fig:cycle_performance}
\end{minipage}
\hfill
\begin{minipage}[t]{0.42\textwidth}
    \vspace{0pt}
    \vspace{2mm}
    \centering
    \captionof{table}{Ablation study of IVAM-Nav on the LiSoNav-Eval subset.}
    \label{tab:ablation}
    \vspace{-3mm}

    \setlength{\tabcolsep}{4pt}
    \renewcommand{\arraystretch}{1.08}
    \resizebox{\linewidth}{!}{%
    \begin{tabular}{lcc}
    \toprule
    Setting & SR$\uparrow$ & SPL$\uparrow$ \\
    \midrule
    w/o cross-task memory
        & 33.06\% & 9.18\% \\
    w/o complementary viewpoints
        & 26.12\% & 13.75\% \\
    w/o graph aggregation
        & 26.53\% & 13.66\% \\
    w/o historical-viewpoint revisit
        & 28.98\% & 14.73\% \\
    w/o memory update
        & 24.90\% & 14.22\% \\
    w/o contextual cues
        & 30.20\% & 15.38\% \\
    \rowcolor{mycolor}
    \textbf{Full}
        & \textbf{36.73\%} & \textbf{21.38\%} \\
    \bottomrule
    \end{tabular}}
\end{minipage}
\vspace{-2mm}
\end{figure*}

\noindent\textbf{Performance across episodes.} Figure~\ref{fig:cycle_performance} reports the first three episodes shared by all sequences, excluding Episode 4 to ensure comparison over the same sequence set. Episode 1 contains no object relocation, while later episodes provide more accumulated memory for reuse but may also introduce stale memory through relocation. IVAM-Nav, with or without a reference image, maintains strong SR across episodes, while its SPL increases after Episode 1 and remains consistently higher thereafter, indicating sustained efficiency benefits from accumulated memory despite potential staleness. In contrast, other cross-task memory methods generally show larger SR degradation or less consistent SPL benefits in later episodes.


\begin{table*}[t!]
\centering
\setlength{\tabcolsep}{8pt}
\renewcommand{\arraystretch}{1.1}
\caption{Performance under static and changing-object-layout settings on the LiSoNav-Eval analysis subset. Relative drops are computed from the static to the changing setting.}
\vspace{-3mm}
\resizebox{0.92\linewidth}{!}{%
\begin{tabular}{lcccccccc}
\toprule
\multirow{2}{*}{Method} &
\multirow{2}{*}{\makecell{Cross-task\\Memory}} &
\multirow{2}{*}{\makecell{Stale-memory\\Revision}} &
\multicolumn{3}{c}{SR} &
\multicolumn{3}{c}{SPL} \\
\cmidrule(lr){4-6}\cmidrule(lr){7-9}
& & &
Static & Changing & Drop (\%) &
Static & Changing & Drop (\%) \\
\midrule
AstraNav~\citep{hu2026astranav}
& Yes & No
& 14.7\% & 11.0\% & 25.2\%
& 12.9\% & 8.5\% & 34.1\% \\
DynaMem~\citep{liu2025dynamem}
& Yes & Yes
& 20.8\% & 17.6\% & 15.7\%
& 15.7\% & 12.6\% & 19.7\% \\
\rowcolor{mycolor}
\textbf{IVAM-Nav}
& Yes & Yes
& \textbf{40.8\%} & \textbf{36.7\%} & \textbf{10.0\%}
& \textbf{25.6\%} & \textbf{21.4\%} & \textbf{16.4\%} \\
\bottomrule
\end{tabular}}
\label{tab:relocation_impact}
\vspace{-3mm}
\end{table*}

\begin{table*}[t]
\vspace{-2mm}
\centering

\begin{minipage}[t]{0.48\textwidth}
\vspace{0pt}
\centering
\captionof{table}{Results by object size. Small and Large correspond to bounding-box
volumes of $<1$\,L and $\geq1$\,L, respectively.}
\label{tab:results_by_object_size}
\vspace{-3mm}

\setlength{\tabcolsep}{7pt}
\renewcommand{\arraystretch}{1.1}
\resizebox{\linewidth}{!}{%
\begin{tabular}{lcccc}
\toprule
\multirow{2}{*}{Method} &
\multicolumn{2}{c}{Small} &
\multicolumn{2}{c}{Large} \\
\cmidrule(lr){2-3}\cmidrule(lr){4-5}
& SR$\uparrow$ & SPL$\uparrow$
& SR$\uparrow$ & SPL$\uparrow$ \\
\midrule
VLFM~\citep{yokoyama2024vlfm}
& 11.2\% & 7.4\%
& 14.8\% & 12.0\% \\

AstraNav~\citep{hu2026astranav}
& 12.5\% & 9.6\%
& 11.5\% & 9.1\% \\

DualMap~\citep{jiang2025dualmap}
& 13.2\% & 9.5\%
& 15.6\% & 11.8\% \\

DynaMem~\citep{liu2025dynamem}
& 17.9\% & 12.7\%
& 18.3\% & 12.8\% \\

3D-Mem~\citep{yang2025threedmem}
& 20.9\% & 14.4\%
& 24.3\% & 18.1\% \\

IVAM-Nav
& 29.2\% & 16.4\%
& 42.9\% & 23.0\% \\

IVAM-Nav (w/ reference image)
& 33.3\% & 19.5\%
& 43.2\% & 25.4\% \\
\bottomrule
\end{tabular}}
\end{minipage}
\hfill
\begin{minipage}[t]{0.48\textwidth}
\vspace{0pt}
\centering
\captionof{table}{Results by relocation distance. Short and Long correspond
to geodesic displacements of $<7.5$\,m and $\geq7.5$\,m, respectively.}
\label{tab:results_by_movement_distance}
\vspace{-3mm}

\setlength{\tabcolsep}{7pt}
\renewcommand{\arraystretch}{1.1}
\resizebox{\linewidth}{!}{%
\begin{tabular}{lcccc}
\toprule
\multirow{2}{*}{Method} &
\multicolumn{2}{c}{Short} &
\multicolumn{2}{c}{Long} \\
\cmidrule(lr){2-3}\cmidrule(lr){4-5}
& SR$\uparrow$ & SPL$\uparrow$
& SR$\uparrow$ & SPL$\uparrow$ \\
\midrule
VLFM~\citep{yokoyama2024vlfm}
& 13.9\% & 10.1\%
& 9.7\% & 7.2\% \\

AstraNav~\citep{hu2026astranav}
& 11.3\% & 9.1\%
& 7.2\% & 7.2\% \\

DualMap~\citep{jiang2025dualmap}
& 10.5\% & 7.7\%
& 10.1\% & 7.6\% \\

DynaMem~\citep{liu2025dynamem}
& 17.6\% & 11.9\%
& 12.2\% & 7.7\% \\

3D-Mem~\citep{yang2025threedmem}
& 21.0\% & 17.4\%
& 13.6\% & 10.0\% \\

IVAM-Nav 
& 35.5\% & 18.4\%
& 25.5\% & 13.4\% \\

IVAM-Nav (w/ reference image)
& 34.5\% & 18.0\%
& 25.9\% & 13.2\% \\
\bottomrule
\end{tabular}}
\end{minipage}
\vspace{-2mm}
\end{table*}

\begin{table*}[t]
\centering

\begin{minipage}[t]{0.58\textwidth}
\vspace{0pt}
\centering
\captionof{table}{Results by environment scale. Small, Medium, and Large
correspond to navigable areas of $<32$\,m$^2$, $32$--$<64$\,m$^2$, and
$\geq64$\,m$^2$, respectively.}
\label{tab:results_by_floor_area}
\vspace{-3mm}

\setlength{\tabcolsep}{7pt}
\renewcommand{\arraystretch}{1.1}
\resizebox{\linewidth}{!}{%
\begin{tabular}{lcccccc}
\toprule
\multirow{2}{*}{Method} &
\multicolumn{2}{c}{Small} &
\multicolumn{2}{c}{Medium} &
\multicolumn{2}{c}{Large} \\
\cmidrule(lr){2-3}\cmidrule(lr){4-5}\cmidrule(lr){6-7}
& SR$\uparrow$ & SPL$\uparrow$
& SR$\uparrow$ & SPL$\uparrow$
& SR$\uparrow$ & SPL$\uparrow$ \\
\midrule
VLFM~\citep{yokoyama2024vlfm}
& 16.7\% & 12.3\%
& 11.7\% & 8.0\%
& 8.9\% & 6.7\% \\

AstraNav~\citep{hu2026astranav}
& 13.3\% & 11.6\%
& 12.8\% & 10.0\%
& 10.1\% & 6.6\% \\

DualMap~\citep{jiang2025dualmap}
& 18.3\% & 12.1\%
& 13.2\% & 10.2\%
& 11.0\% & 8.4\% \\

DynaMem~\citep{liu2025dynamem}
& 20.7\% & 15.6\%
& 18.5\% & 12.9\%
& 14.6\% & 9.6\% \\

3D-Mem~\citep{yang2025threedmem}
& 31.6\% & 26.1\%
& 19.6\% & 12.8\%
& 16.4\% & 9.9\% \\

IVAM-Nav
& 41.5\% & 22.4\%
& 33.7\% & 19.1\%
& 25.0\% & 13.5\% \\

IVAM-Nav (w/ reference image)
& 50.5\% & 30.3\%
& 34.3\% & 20.3\%
& 25.9\% & 14.2\% \\
\bottomrule
\end{tabular}}
\end{minipage}
\hfill
\begin{minipage}[t]{0.38\textwidth}
\vspace{0pt}
\vspace{1.5mm}
\centering
\captionof{table}{Prior-observation rate of relocated targets.}
\label{tab:moved_target_prior_observation}
\vspace{-3mm}

\setlength{\tabcolsep}{15pt}
\renewcommand{\arraystretch}{1.1}
\resizebox{\linewidth}{!}{%
\begin{tabular}{lc}
\toprule
Method & Rate \\
\midrule
VLFM~\citep{yokoyama2024vlfm}
& 42.4\% \\

AstraNav~\citep{hu2026astranav}
& 55.1\% \\

DynaMem~\citep{liu2025dynamem}
& 49.7\% \\

DynaMem Oracle~\citep{liu2025dynamem}
& 49.4\% \\

3D-Mem~\citep{yang2025threedmem}
& 72.7\% \\

3D-Mem Oracle~\citep{yang2025threedmem}
& 73.8\% \\

IVAM-Nav 
& 71.0\% \\

IVAM-Nav (w/ reference image)
& 74.1\% \\
\bottomrule
\end{tabular}}
\end{minipage}
\vspace{-5mm}
\end{table*}

\vspace{-1mm}
\subsection{Method Analysis}
\label{sec:method_analysis}
\vspace{-2mm}
To enable efficient controlled analysis, we conduct the following experiments on a representative LiSoNav-Eval subset containing 245 subtasks, using the reference-image variant of IVAM-Nav to reduce target-recognition ambiguity and focus the analysis on its memory mechanisms. The representativeness of the subset is analyzed in Appendix, and its composition will be released.

\noindent\textbf{Ablation study.} We consider six variants. \emph{w/o memory} removes cross-task memory entirely. \emph{w/o complementary viewpoints} retains only one viewpoint near each supporting surface during memory construction. \emph{w/o graph aggregation} matches memory candidates independently with the target object without aggregating graph information around historical viewpoints. \emph{w/o historical-viewpoint revisit} navigates to the associated supporting surface for revalidation but does not return to the stored viewpoint. \emph{w/o memory update} leaves inconsistent historical memory unchanged and continues to use it for subsequent navigation decisions. Finally, \emph{w/o contextual cues} removes the surface and surroundings cues from the instruction, using only the small-object description for memory retrieval, to evaluate IVAM-Nav when such contextual information is unavailable.

As shown in Table~\ref{tab:ablation}, removing cross-task memory reduces SPL from 21.38\% to 9.18\%, showing the benefit of accumulated memory for navigation efficiency. Retaining only a single viewpoint for each surface reduces SR from 36.73\% to 26.12\%, supporting the use of complementary viewpoints to obtain reliable small-object information. Without graph aggregation, SR decreases to 26.53\%, showing that aggregating scene information around historical viewpoints improves memory retrieval over independently matching memory candidates.
Without historical-viewpoint revisit for revalidation, SR drops to 28.98\%, showing the benefit of revisiting the original observation viewpoint under similar viewing conditions. Disabling memory update yields the largest SR drop, to 24.90\%, highlighting the importance of revising stale memory rather than continuing to use it for subsequent decisions. 

Finally, even without contextual cues, IVAM-Nav retains an SR of 30.20\%, indicating that the method remains effective when it relies only on the small-object description.

\noindent\textbf{Robustness to object relocation.} To assess how robustly different methods handle object relocation and the resulting stale memory, we compare representative cross-task memory methods under static and changing-object-layout settings. The static setting removes cross-episode object relocation, with details provided in Appendix. As shown in Table~\ref{tab:relocation_impact}, IVAM-Nav exhibits the smallest degradation under scene changes, with relative SR and SPL drops of 10.0\% and 16.4\%, respectively. This indicates that IVAM-Nav is less affected by object relocation and the potentially stale memory it introduces. See Appendix for further analysis.

\subsection{Benchmark Analysis}
\label{sec:benchmark_analysis}
\vspace{-2mm}
We analyze LiSoNav-Eval along factors that affect small-object perception, target search, and adaptation to scene changes. As shown in Table~\ref{tab:results_by_object_size}, most methods perform worse on smaller objects, reflecting the increased difficulty of acquiring reliable visual information when targets occupy a limited image area. Table~\ref{tab:results_by_floor_area} shows a clear performance decline as environment scale increases, indicating the greater exploration and search difficulty in larger spaces. For relocated targets, Table~\ref{tab:results_by_movement_distance} shows that most methods perform worse under longer relocation distances, suggesting that larger location changes make recovery from outdated target memory more challenging.

\noindent\textbf{Prior target observation.}
To assess whether relocation may invalidate acquired target information, we
measure the fraction of relocated targets observed before relocation using
ground-truth simulator information and the HR observation criterion. As shown in
Table~\ref{tab:moved_target_prior_observation}, prior-observation rates range
from 42.4\% to 74.1\% across methods, showing that relocation can frequently
make historical target information outdated. Moreover, the SR gap between
unchanged and relocated targets is larger for previously observed targets
than for unobserved targets for both IVAM-Nav (12.08 vs.\ 3.11 percentage
points) and DynaMem (9.00 vs.\ 5.84 percentage points). See for details.

\vspace{-1mm}
\section{Limitations and Future Directions}
\vspace{-2mm}
IVAM-Nav leaves several directions for future work. Active multi-view observation introduces additional navigation cost, motivating more cost-aware acquisition strategies. Memory revalidation is mainly triggered by retrieved historical viewpoints, motivating broader and opportunistic memory maintenance. Memory reuse and exploration are currently sequential and could be more tightly integrated. Finally, LiSoNav-Eval relocates small objects while keeping background scenes stable; future benchmarks could consider broader scene changes. See Appendix for further discussion.

\vspace{-1mm}
\section{Conclusion}
\vspace{-2mm}
We introduced LiSoNav-COL, which studies lifelong small-object navigation from empty memory under changing object layouts, together with the LiSoNav-Eval benchmark. We further proposed IVAM-Nav, which combines multi-view inspection for reliable memory construction with viewpoint-anchored memory reuse and revalidation. Experiments show consistent improvements over existing navigation and scene-memory methods and validate the key designs of IVAM-Nav.

\bibliography{iclr2027_conference}
\bibliographystyle{iclr2027_conference}
\clearpage

\end{document}